\documentclass[conference,letterpaper]{IEEEtran}
\IEEEoverridecommandlockouts

\ifCLASSINFOpdf
\else
\fi

\usepackage[dvips]{graphicx}
\usepackage{amsmath,amssymb}
\usepackage{paralist}
\usepackage{subfigure}
\usepackage{url}
\usepackage{algorithm}
\usepackage{algorithmic}

\makeatletter
\def\tbcaption{\def\@captype{table}\caption}
\def\figcaption{\def\@captype{figure}\caption}

\newcommand{\bvec}[1]{\mbox{\boldmath $#1$}}

\begin{document}
\title{Knowledge Extracted from Recurrent Deep Belief Network for Real Time Deterministic Control
\thanks{\copyright 2017 IEEE. Personal use of this material is permitted. Permission from IEEE must be obtained for all other uses, in any current or future media, including reprinting/republishing this material for advertising or promotional purposes, creating new collective works, for resale or redistribution to servers or lists, or reuse of any copyrighted component of this work in other works.}
} 

\author{
\IEEEauthorblockN{Shin Kamada}
\IEEEauthorblockA{Graduate School of Information Sciences, \\
Hiroshima City University\\
3-4-1, Ozuka-Higashi, Asa-Minami-ku,\\
Hiroshima, 731-3194, Japan\\
Email: da65002@e.hiroshima-cu.ac.jp}
\and 
\IEEEauthorblockN{Takumi Ichimura}
\IEEEauthorblockA{Faculty of Management and Information Systems,\\
Prefectural University of Hiroshima\\
1-1-71, Ujina-Higashi, Minami-ku,\\
Hiroshima, 734-8559, Japan\\
Email: ichimura@pu-hiroshima.ac.jp}}

\maketitle

\begin{abstract}
  Recently, the market on deep learning including not only software but also hardware is developing rapidly. Big data is collected through IoT devices and the industry world will analyze them to improve their manufacturing process. Deep Learning has the hierarchical network architecture to represent the complicated features of input patterns. Although deep learning can show the high capability of classification, prediction, and so on, the implementation on GPU devices are required. We may meet the trade-off between the higher precision by deep learning and the higher cost with GPU devices. We can success the knowledge extraction from the trained deep learning with high classification capability. The knowledge that can realize faster inference of pre-trained deep network is extracted as IF-THEN rules from the network signal flow given input data. Some experiment results with benchmark tests for time series data sets showed the effectiveness of our proposed method related to the computational speed.
\end{abstract}

\IEEEpeerreviewmaketitle

\section{Introduction}
Recently, the market on deep learning including not only software but also hardware is developing rapidly. According to the new market research report \cite{webmarket2016}, this market is expected to be worth more than USD 1770 billion by 2022, growing at a CAGR (Compound Annual Growth Rate) of 65.3\% between 2016 and 2022.

Many new algorithms in various learning structures of deep learning have been reported and a new methodology on Artificial Intelligence (AI) has permeated through industry world. In 2016, the IoT (Internet of Things) Acceleration Consortium was established in Japan with the aim of creating an adequate environment for attracting investment in the future with the IoT through public-private collaboration. The collected big data though IoT technologies is analyzed by AI technologies including deep learning. Deep learning technology becomes a popular and sophisticated technology embedded in the measuring machine that the higher precision is required \cite{Bengio09}.

For example, a machine that has a robot arm with closed-loop control should know the precise position data immediately to determine the next move. The speed of the events being controlled is an important and critical factor in a real-time operation. Among events, a machine must respond the determined next position within absolute limit response time to predict the following events in its environment. A machine must realize a strict control between hard and soft real-time necessities.

Deep learning algorithm works the implementation of task on GPU (Graphics Processing Unit) device. In a hard real-time system such as embed system with GPU, the machine need a high cost to build an overall system. But in a soft real-time system, a late response makes an inevitable result such as the deterioration of production efficiency. We may meet the trade-off between the higher precision by deep learning and the higher cost with GPU devices.

Deep Learning has the hierarchical network architecture to represent the complicated features of input patterns \cite{Hinton06}. Such architecture is well known to represent higher learning capability compared with some conventional models if the best set of parameters in the optimal network structure is found. We have been developing the adaptive learning method that can discover the optimal network structure in Deep Belief Network (DBN) \cite{Kamada16_SMC,Kamada16_ICONIP,Kamada16_TENCON}. The learning method can construct the network structure with the optimal number of hidden neurons in each Restricted Boltzmann Machine \cite{Hinton12} and with the optimal number of layers in the DBN during learning phase. Moreover, we develop the recurrent neural network based Deep Belief Network (DBN) to make a higher predictor to the time series data set \cite{Ichimura17_IJCNN}.

However, the implementation of our developed deep learning method required some expensive GPU devices because the computation time to output feed-forward calculation is too long to realize a real-time control in the manufacturing process or image diagnosis device. In order to make the spread of our developed method, the operating method in the small embedded system or smart tablet without GPU devices is required.

In this paper, the knowledge that can realize faster inference of pre-trained deep network is extracted as IF-THEN rules from the network signal flow given input data. Some experiment results with benchmark test for time series data set showed the effectiveness of our proposed method related to the computational speed.

\section{Adaptive Learning Method of RNN-DBN}
\begin{figure}[tbp]
\begin{center}
\includegraphics[scale=0.5]{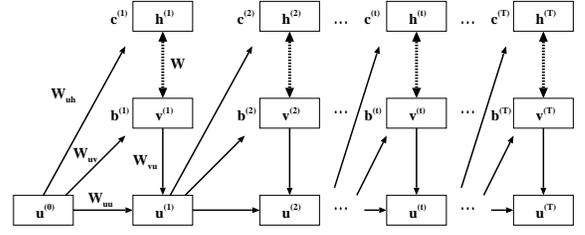}
\caption{Recurrent Neural Network RBM\cite{Lewandowski12}}
\label{fig:rnn-rbm}
\end{center}
\end{figure}

\begin{figure}[tbp]
\begin{center}
\includegraphics[scale=0.5]{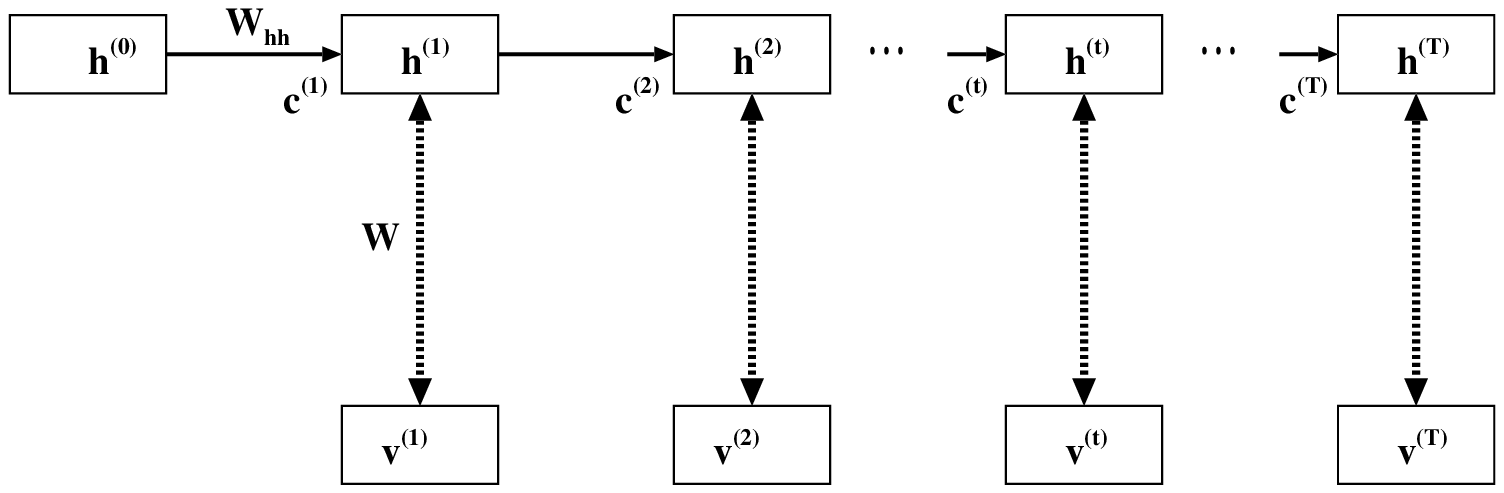}
\caption{Temporal RBM\cite{Sutskever08}}
\label{fig:trbm}
\end{center}
\end{figure}

\begin{figure}[tbp]
\begin{center}
\includegraphics[scale=0.5]{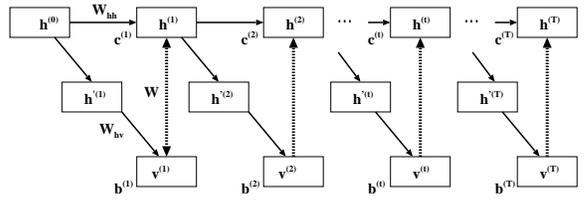}
\caption{Recurrent Temporal RBM\cite{Sutskever08}}
\label{fig:rtrbm}
\end{center}
\end{figure}

Recurrent Neural Network Restricted Boltzmann Machine (RNN-RBM) \cite{Lewandowski12} is known to be a unsupervised learning algorithm for a time series data based on RBM model \cite{Hinton12}. Fig.~\ref{fig:rnn-rbm} shows the network structure of RNN-RBM. The model forms a directed graphical model consisting of a sequence of RBMs such as a recurrent neural network as well as Temporal RBM (Fig.~\ref{fig:trbm}) or Recurrent TRBM (Fig.~\ref{fig:rtrbm}) \cite{Sutskever08}.

The RNN-RBM model has the state of representing contexts in time series data, $\bvec{u} \in \{0, 1 \}^{K}$, related to past sequences of time series data in addition to the visible neurons and the hidden neurons of the traditional RBM. Let the sequence of input sequence with the length $T$ be $\bvec{V} = \{ \bvec{v}^{(1)}, \cdots, \bvec{v}^{(t)}, \cdots, \bvec{v}^{(T)} \}$. The parameters $\bvec{b}^{(t)}$ and $\bvec{c}^{(t)}$ for the visible layer and the hidden layer, respectively are calculated from $\bvec{u}^{(t-1)}$ for time $t-1$ by using Eq.(\ref{eq:calc_bt}) and Eq.(\ref{eq:calc_ct}). The state $\bvec{u}^{(t)}$ at time $t$ is updated by using Eq.(\ref{eq:calc_ut}).
\begin{equation}
\label{eq:calc_bt}
\bvec{b}^{(t)} = \bvec{b} + \bvec{W}_{uv} \bvec{u}^{(t-1)},
\end{equation}
\begin{equation}
\label{eq:calc_ct}
\bvec{c}^{(t)} = \bvec{c} + \bvec{W}_{uh} \bvec{u}^{(t-1)},
\end{equation}
\begin{equation}
\label{eq:calc_ut}
\bvec{u}^{(t)} = \sigma( \bvec{u} + \bvec{W}_{uu} \bvec{u}^{(t-1)} + \bvec{W}_{vu} \bvec{v}^{(t)} ),
\end{equation}
where $\sigma()$ is a sigmoid function. $\bvec{{u}}^{(0)}$ is the initial state which is given a random value. At each time $t$, the learning of traditional RBM can be executed with $\bvec{b}^{(t)}$ and $\bvec{c}^{(t)}$ at time $t$ and weights $\bvec{W}$ between them. After the error are calculated till time $T$, the gradients for $\bvec{\theta}=\{\bvec{b}, \bvec{c}, \bvec{W}, \bvec{u}, \bvec{W}_{uv}, \bvec{W}_{uh}, \bvec{W}_{vu}, \bvec{W}_{uu} \}$ are updated to trace from time $T$ back to time $t$ by BPTT (Back Propagation Through Time) method \cite{Elman90,Jordan86}.

\begin{figure}[tbp]
\begin{center}
\subfigure[Neuron Generation]{\includegraphics[scale=0.5]{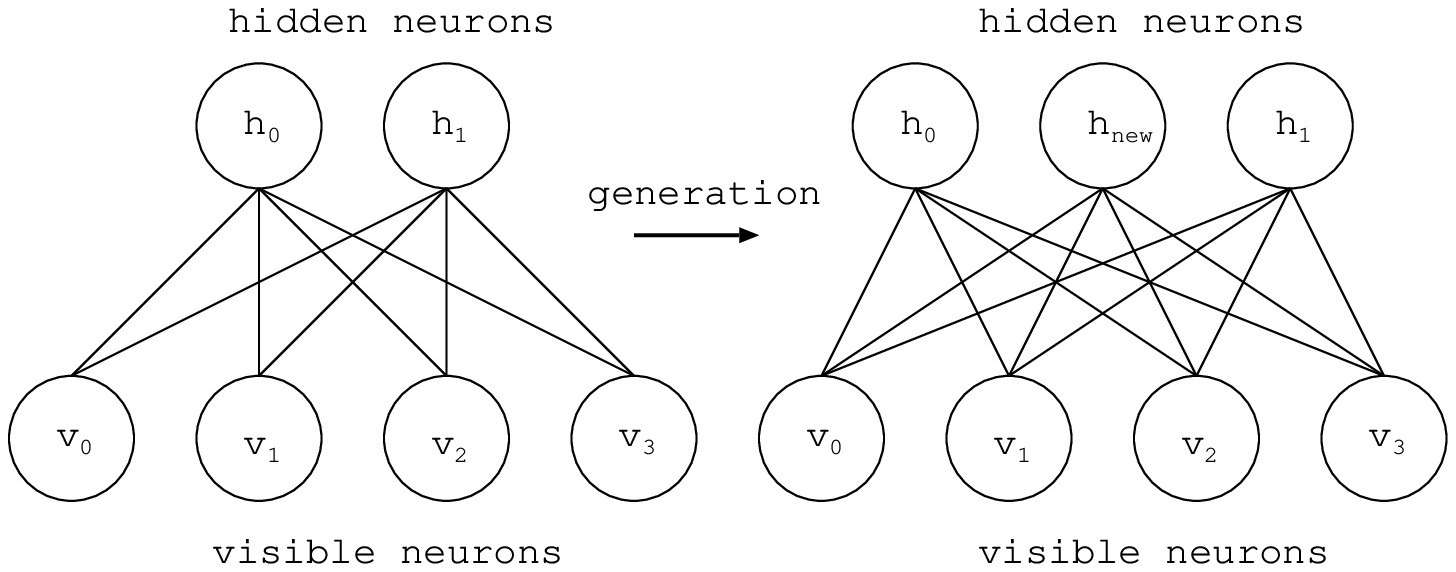}
  \label{fig:neuron_generation}
}
\subfigure[Neuron Annihilation]{\includegraphics[scale=0.5]{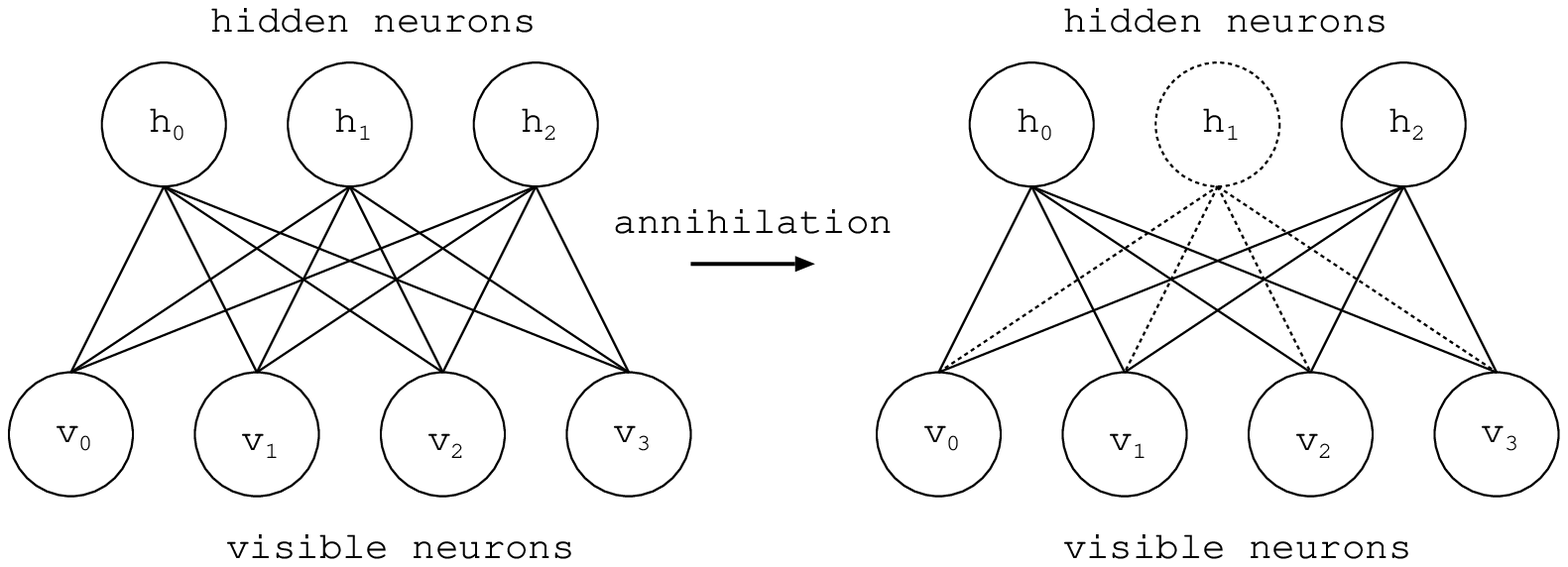}
  \label{fig:neuron_annihilation}
}
\vspace{-3mm}
\caption{Adaptive RBM}
\label{fig:adaptive_rbm}
\vspace{-3mm}
\end{center}
\end{figure}

\begin{figure}[tbp]
\begin{center}
\includegraphics[scale=0.6]{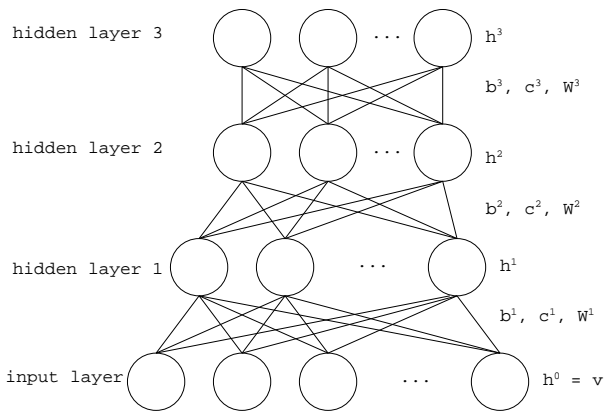}
\caption{An overview of DBN}
\label{fig:dbn}
\end{center}
\end{figure}

\begin{figure*}[tbp]
\begin{center}
\includegraphics[scale=0.8]{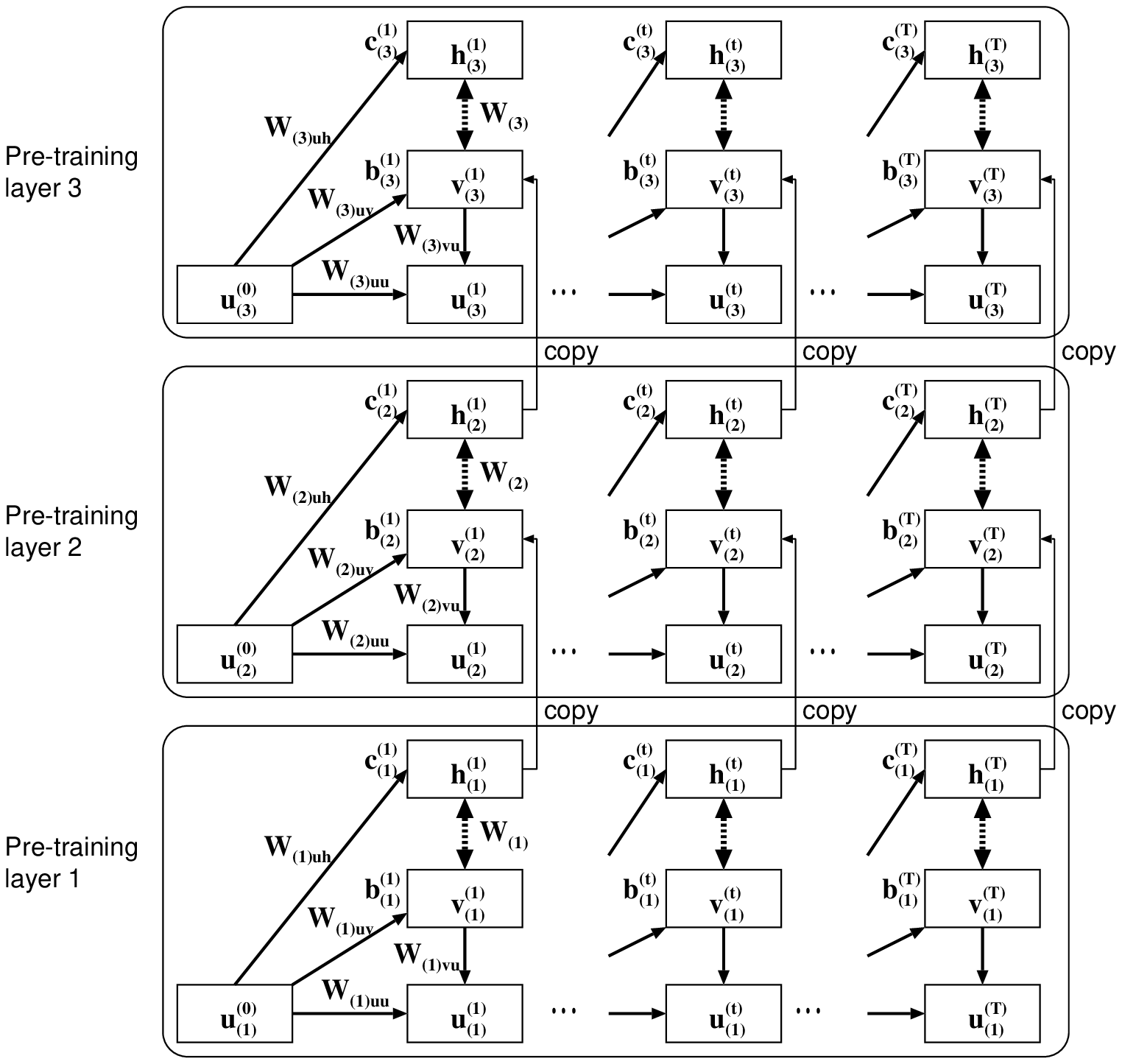}
\caption{Recurrent Neural Network DBN}
\label{fig:rnn-dbn}
\end{center}
\end{figure*}

We proposed the adaptive learning method of RNN-RBM with self-organization function of network structure according to a given input data \cite{Ichimura17_IJCNN}. The optimal number of hidden neurons can be automatically determined according to the variance of weight and parameter of the hidden neurons during the learning by neuron generation / annihilation algorithms as shown in Fig.\ref{fig:adaptive_rbm} \cite{Kamada16_SMC,Kamada16_ICONIP}. The structure of general deep belief network as shown in Fig.\ref{fig:dbn} is the accumulating 2 or more RBMs. For RNN-RBM, an ingenious contrivance is required to treat time series data. RNN-RBM was improved by building the hierarchical network structure to pile up the pre-trained RNN-RBM. As shown in Fig.~\ref{fig:rnn-dbn}, the output signal of hidden neuron $\bvec{h}^{(t)}$ at time $t$ can be seen as the input signal of visible neuron $\bvec{v}^{(t)}$ of the next layer of RNN-RBM. We also proposed the adaptive learning method of RNN-DBN that can determine the optimal number of hidden layers for a given input data \cite{Kamada16_TENCON}.

\section{Knowledge Discovery}
The knowledge discovery is one of the approaches that we should solve in Deep Learning. The inference with the trained deep neural network is considered to realize highly classification capability. However, the network trained by the deep neural network forms a black box. It is difficult for us to interpret the inference process of Deep Learning intuitively because the trained network has just a numerical value for weight or outputs of hidden neurons. In addition, sometimes hardware with highly performance or the GPU embedded devices may be required when utilizing the trained network, in terms of calculation speed. Therefore, a method to make a sparse network structure from the trained network or to extract the knowledge with respect to the inference process of deep neural network as IF-THEN rules will be a helpful solution. Ishikawa proposed the structural learning method to make a sparse network structure in multi-layered neural networks \cite{Ishikawa96}. The method of knowledge discovery by using the distillation model was proposed by Hinton \cite{Hinton15}. The distillation means to transfer the knowledge from the complicated model to a small model.

In our recent research, the fine-tuning method with the pre-trained network was developed \cite{Kamada16_IWCIA}. The basic idea of the approach is to calculate the frequency of fired hidden neurons for a given input and to find the path of network signal flows from input layer to output layer. After then, a wrong signal flow which is related to incorrect output will be fine-tuned correctly. By using the method, the improvement of classification capability was verified by using experimental results for some benchmark data sets. However, the method is not a kind of knowledge discovery to extract the knowledge of inference process of deep neural network.

In this paper, we propose another method of knowledge discovery by using C4.5 \cite{Quinlan96}, that the extracted knowledge can realize faster inference process instead of the pre-trained network. C4.5 is a well known and popular method to generate a decision tree, although fuzzy based decision tree method has developed. Because the C4.5 is easily applied in the industrial products, but fuzzy based decision tree remains some difficulties in the determination of fuzzy sets so that the classifier to control fast super machine based on fuzzy theory is not used. Therefore, we use C4.5 method to make decision tree for knowledge discover from trained neural network.

For the training data, the teach signal to the input pattern is given and the relation between input and output pattern are extracted by C4.5. However, the teach signal to the input data are not given and the output pattern are determined by the feed forward calculation. When the knowledge is extracted from the trained network, the pairwise data of input pattern and the calculated output pattern are analyzed. The decision trees generated by C4.5 can be used for classification of pairwise data. 

\begin{figure}[bp]
\centering
\includegraphics[scale=0.6]{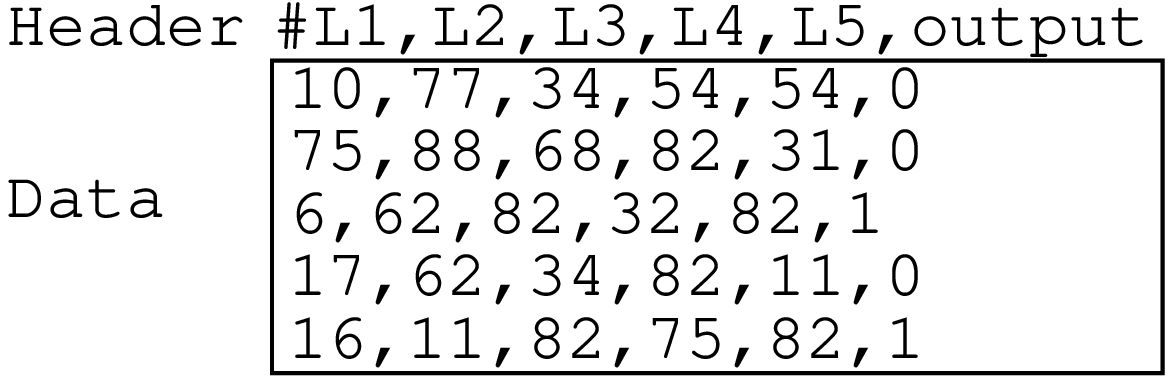}
\caption{An example of C4.5 data file}
\label{fig:c45_sample}
\end{figure}

The procedure of knowledge discovery is as follows. First, the hidden neuron that is the most fired hidden neuron at each layer is calculated for a given input at time $t$. Next, the data file to analysis by C4.5 is created. The input is the network path from input to output layer, and output is the teacher signal, that is output value at $t+1$. Fig.~\ref{fig:c45_sample} shows an example of C4.5 data file. The numerical value in Fig.~\ref{fig:c45_sample} means an index of a hidden neuron. For example, the first line in Fig.~\ref{fig:c45_sample} means that the neurons are fired through the path `10' $\rightarrow$ `77' $\rightarrow$ `34' $\rightarrow$ `54' $\rightarrow$ `54' from 1st layer to 5th layer, and the output value is `0'.

\begin{table*}[tbp]
\caption{Prediction Accuracy on Nottingham}
\vspace{-5mm}
\label{tab:accuracy_nottingham}
\begin{center}
\begin{tabular}{l|r|r|r|r}
\hline \hline
& No. layers & Error(Training) & Error(Test) & Accuracy (Test, \%) \\ \hline\hline
Traditional RNN-RBM              & 1   &  0.945  &  1.704  &  71.7\% \\ \hline
Adaptive  RNN-RBM                & 1   &  {\bf 0.881}  &  {\bf 1.240}  &  {\bf 76.5}\% \\ \hline \hline
Traditional RNN-DBN              & 5   &  0.217    & 1.381 &  75.8\% \\ \hline 
Adaptive  RNN-DBN                & 5   &  {\bf 0.101}    &  {\bf 0.133} &  {\bf 93.9}\% \\ \hline
\hline 
\end{tabular}
\end{center}
\end{table*}

\begin{table*}[htb]
\caption{Prediction Accuracy on CMU}
\vspace{-5mm}
\label{tab:accuracy_cmu}
\begin{center}
\begin{tabular}{l|r|r|r|r}
\hline \hline
& No. layers & Error(Training) & Error(Test) & Correct ratio (Test) \\ \hline\hline
Traditional RNN-RBM              & 1   &  1.344  &  2.401  &  65.2\% \\ \hline
Adaptive  RNN-RBM                & 1   &  {\bf 0.981}  &  {\bf 1.471}  &  {\bf 73.1}\% \\ \hline \hline
Traditional RNN-DBN              & 6   &  0.517    & 2.202 &  70.8\% \\ \hline 
Adaptive  RNN-DBN                & 6   &  {\bf 0.121}    &  {\bf 0.148} &  {\bf 82.3}\% \\ \hline
\hline 
\end{tabular}
\end{center}
\end{table*}

\section{Experimental Results for Benchmark Data Set}
\subsection{Data Sets}
\begin{figure}[tbp]
\begin{center}
\includegraphics[scale=1.3]{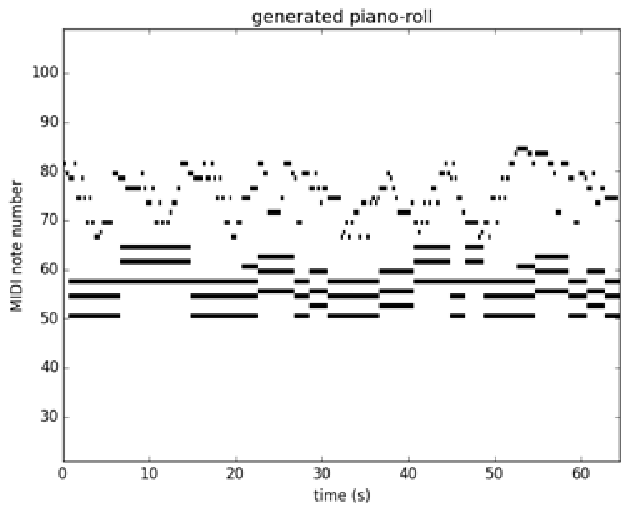}
\caption{An example of Nottingham}
\label{fig:Nottingham_example}
\end{center}
\end{figure}

\begin{figure}[tbp]
\begin{center}
\includegraphics[scale=0.5]{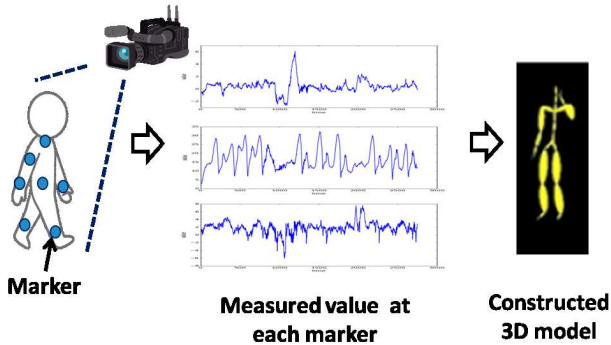}
\caption{An example of CMU}
\label{fig:CMU_example}
\end{center}
\end{figure}

In the simulation described here, the benchmark data set `Nottingham' \cite{Nottingham} and `CMU' \cite{CMU} were used to verify the effectiveness of our proposed method. Nottingham is a classical piano MIDI archive included in about 694 training cases and about 170 test cases. Fig.~\ref{fig:Nottingham_example} is an example of MIDI. Each MIDI includes sequential data for about 60 seconds. The horizontal axis in Fig.~\ref{fig:Nottingham_example} shows the time (second) and the vertical axis shows the MIDI note number from 0 to 87. Therefore an input data at time $t$ is represented as 88 dimensional $\{0, 1\}$ vector, where `1' means the MIDI key is stroked, `0' is opposite. On the other hand, CMU is a motion capture database collected by Carnegie Mellon University. There are 2,605 trials in 6 categories which are divided into 23 subcategories. The data represents a human movement for about 30 seconds attached with more than 30 markers. The 3D model can be constructed from the measured value by the markers as shown in Fig.~\ref{fig:CMU_example}.

The parameters used in this paper are as follows. The training algorithm is Stochastic Gradient Descent (SGD) method, the batch size is 100, and the learning rate is 0.001. In the experiment, 2 kinds of computers are used. One is a high-end GPU embedded workstation for the training and the knowledge discovery. The specification is as follows. CPU: Intel(R) 24 Core Xeon E5-2670 v3 2.3GHz, GPU: Tesla K80 4992 24GB $\times$ 3, Memory: 64GB, and OS: Cent OS 6.7 64 bit. The other computer is a low-end machine such as embedded system in the industrial world. The prediction accuracy and its calculation speed are evaluated by using the pre-trained network on GPU and the acquired knowledge without GPU in this computer. The specification is as follows. CPU: Intel(R) Core(TM) i5-4460 @ 3.20GHz, GPU: GTX 1080 8GB, Memory: 8GB, and OS: Fedora 23 64 bit.

\begin{figure}[tbp]
\centering
\includegraphics[scale=0.8]{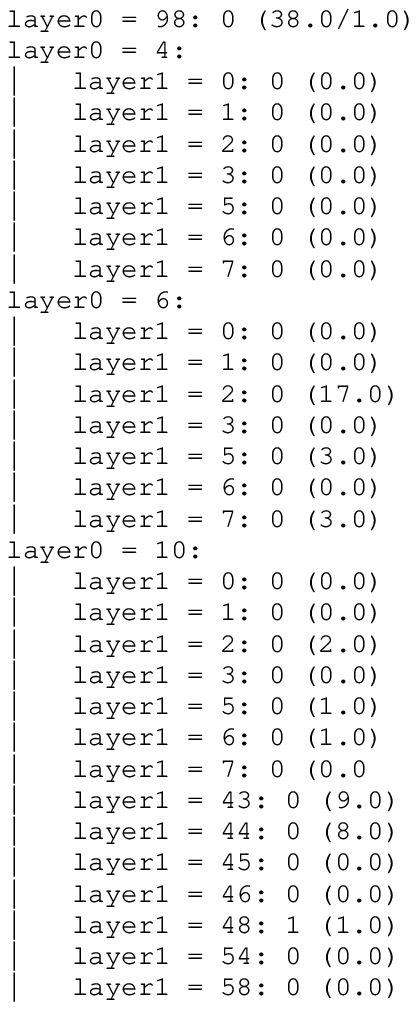}
\vspace{-3mm}
\caption{Sample of Acquired Tree by C4.5}
\label{fig:c45_tree}
\end{figure}

\subsection{Experimental Results}
Table\ref{tab:accuracy_nottingham} and Table\ref{tab:accuracy_cmu} show the prediction accuracy on Nottingham and CMU, respectively. Our proposed Adaptive RNN-RBM can obtain smaller error and higher prediction accuracy than the traditional RNN-RBM for not only the training case but also the test case. Moreover, Adaptive RNN-DBN can improve the prediction accuracy greatly because 4 additional layers were automatically generated by the neuron generation / annihilation algorithms and the layer generation condition. The traditional RNN-DBN does not have the mechanism of neuron generation and the network is not trained under the optimal hidden neurons. The DBN learning piled the layer up no optimized structure of lower layer. Our method with neuron generation finds the optimal structure and then it can perform higher classification capability than the traditional method.

The knowledge was extracted from the trained Adaptive RNN-DBN by C4.5. The number of extracted rules is 125 and 153 for Nottingham and CMU, respectively. Fig.~\ref{fig:c45_tree} shows the generated tree structure by C4.5. By applying the rules into the inference process instead of using the trained network, the prediction results as shown in Table~\ref{tab:accuracy_nottingham_knowledge} and Table~\ref{tab:accuracy_cmu_knowledge} are acquired for Nottingham and CMU, respectively. `Without knowledge' shows the calculation result on GPU with the trained Adaptive RNN-DBN network. `With knowledge' is the result with the acquired knowledge instead of the trained network. Although the prediction accuracy with the knowledge was slight lower than the result without the knowledge, the proposed method with the knowledge obtained highly computational speed. Especially, the proposed method improved CPU time to about 1/40 even if not using the GPU board.

\begin{table}[tbp]
\caption{Prediction Accuracy on Nottingham with Knowledge}
\vspace{-5mm}
\label{tab:accuracy_nottingham_knowledge}
\begin{center}
\begin{tabular}{l|r|r}
\hline \hline
& Accuracy (\%) & CPU time(s) \\ \hline\hline
Without knowledge    & 93.9 & 0.85 \\ \hline
With knowledge       & 90.1 & 0.02 \\ \hline
\hline 
\end{tabular}
\end{center}
\end{table}
\begin{table}[tbp]
\caption{Prediction Accuracy on CMU with Knowledge}
\vspace{-5mm}
\label{tab:accuracy_cmu_knowledge}
\begin{center}
\begin{tabular}{l|r|r}
\hline \hline
& Accuracy (\%) & CPU time(s) \\ \hline\hline
Without knowledge   & 82.3 & 0.38 \\ \hline
With knowledge      & 80.9 & 0.01 \\ \hline
\hline 
\end{tabular} 
\end{center}
\end{table}

\section{Conclusive Discussion}
Deep Learning is widely used in various kinds of research fields, especially image recognition. Although the deep network structure has high classification capability, the problem is that the computational cost with GPU boards should be required when training network and testing with the trained one. In our research, the adaptive learning method of DBN that can self-organize the optimal network structure according to the given input data during the learning phase has been developed. Our proposed adaptive learning method of DBN records a great score as for not only image classification task but also prediction task of time-series data set. Moreover, the method of knowledge discovery to extract the rules which predict a next status of time series data from the trained network was proposed in this paper. The experimental results showed that the extracted knowledge realized faster prediction speed without using the trained deep neural network. However, the trade-off between precise of reasoning and the computation time is still remained. We will make a solution for this problem in future work. 

\section*{Acknowledgment}
This work was supported by JAPAN MIC SCOPE Grand Number 162308002, Artificial Intelligence Research Promotion Foundation, and JSPS KAKENHI Grant Number JP17J11178.


\begin{thebibliography}{1}

\bibitem{webmarket2016}
Markets and Markets, \url{http://www.marketsandmarkets.com/Market-Reports/deep-learning-market-107369271.html} [online] (2016)
  
\bibitem{Bengio09}
Y.Bengio, \emph{Learning Deep Architectures for AI}. Foundations and Trends in Machine Learning archive, Vol.2, No.1, pp.1--127 (2009)

\bibitem{Hinton06}
G.E.Hinton, S.Osindero and Y.Teh, \emph{A fast learning algorithm for deep belief nets}. Neural Computation, Vol.18, No.7, pp.1527--1554 (2006)
  
\bibitem{Kamada16_SMC}
S.Kamada and T.Ichimura, \emph{An Adaptive Learning Method of Restricted Boltzmann Machine by Neuron Generation and Annihilation Algorithm}. Proc. of 2016 IEEE SMC (SMC2016), pp.1273--1278 (2016)

\bibitem{Kamada16_ICONIP}
S.Kamada, T.Ichimura, \emph{A Structural Learning Method of Restricted Boltzmann Machine by Neuron Generation and Annihilation Algorithm}, Neural Information Processing, Proc. of the 23rd International Conference on Neural Information Processing, Springer LNCS9950), pp.372--380 (2016)

\bibitem{Kamada16_TENCON}
S.Kamada and T.Ichimura, \emph{An Adaptive Learning Method of Deep Belief Network by Layer Generation Algorithm}, Proc. of IEEE TENCON2016, pp.2971--2974 (2016)
  
\bibitem{Hinton12}
G.E.Hinton, \emph{A Practical Guide to Training Restricted Boltzmann Machines}. Neural Networks, Tricks of the Trade, Lecture Notes in Computer Science, Vol.7700, pp.599--619 (2012)

\bibitem{Ichimura17_IJCNN} 
T.Ichimura, S.Kamada, \emph{Adaptive Learning Method of Recurrent Temporal Deep Belief Network to Analyze Time Series Data}, Proc. of the 2017 International Joint Conference on Neural Network (IJCNN 2017), pp.2346--2353 (2017)

\bibitem{Lewandowski12}
N.B.Lewandowski, Y.Bengio and P.Vincent,
\emph{Modeling Temporal Dependencies in High-Dimensional Sequences:Application to Polyphonic Music Generation and Transcription}, Proc. of the 29th International Conference on Machine Learning (ICML 2012), pp.1159--1166 (2012)

\bibitem{Sutskever08}
I.Sutskever, G.E.Hinton, and G.W.Taylor,
\emph{The Recurrent Temporal Restricted Boltzmann Machine}, Proc. of Advances in Neural Information Processing Systems 21 (NIPS-2008) (2008)

\bibitem{Elman90}
J.Elman, \emph{Finding structure in time}, Cognitive Science, Vol.14, No.2 (1990)

\bibitem{Jordan86}
  M.Jordan, \emph{Serial order: A parallel distributed processing approach}, Tech. Rep. No. 8604. San Diego: University of California, Institute for Cognitive Science (1986)

\bibitem{Ishikawa96}
M.Ishikawa, \emph{Structural Learning with Forgetting}, Neural Networks, Vol.9, No.3, pp.509--521 (1996)

\bibitem{Hinton15}
G.Hinton, O.Vinyals, J.Dean, \emph{Distilling the Knowledge in a Neural Network}, Proc. of NIPS 2014 Deep Learning Workshop (2014)

\bibitem{Kamada16_IWCIA}
S.Kamada and T.Ichimura, \emph{Fine Tuning Method by using Knowledge Acquisition from Deep Belief Network}, Proc. of IEEE 9th IWCIA2016, pp.119--124 (2016)

\bibitem{Quinlan96}
  J.R.Quinlan, \emph{Improved use of continuous attributes in c4.5}, Journal of Artificial Intelligence Research, No.4, pp.77--90 (1996)

\bibitem{Nottingham}
Nottingham, \url{http://www-etud.iro.umontreal.ca/~boulanni/icml2012} [online] (2016)
  
\bibitem{CMU}
CMU Graphics Lab Motion Capture Database, \url{http://mocap.cs.cmu.edu/} [online] (2016)

\end{thebibliography}
\end{document}